\documentclass[runningheads]{llncs}
\usepackage[T1]{fontenc}
\usepackage{graphicx}
\usepackage{amsmath,amssymb,amsfonts}
\usepackage{mathtools}

\DeclareMathOperator{\Perc}{Perc}

\usepackage{multirow}
\usepackage{makecell}

\begin{document}
\title{Raymoval: Raycasting-based Dynamic Object Removal for Static 3D Mapping}
\titlerunning{Raymoval: Raycasting-based Dynamic Object Removal}

\author{Daebeom Kim \and
Seungjae Lee \and Seoyeon Jang \and Kevin Christiansen Marsim \and 
Hyun Myung}
\authorrunning{D. Kim et al.}

\institute{
School of Electrical Engineering, KAIST (Korea Advanced Institute of Science and
Technology), Daejeon 34141, Republic of Korea\\
\email{\{ted97k, sj98lee, 9uantum01, kevinmarsim, hmyung\}@kaist.ac.kr}\\
\url{http://urobot.kaist.ac.kr}}
\maketitle              
\begin{abstract}
Static mapping is fundamental to robot navigation, providing a persistent geometric prior and a consistent reference for long-term autonomy. However, dynamic objects leave residual traces and cause surface loss, which reduces map consistency. We propose a raycasting-based module for dynamic object removal in static 3D mapping. Each scan is projected onto an azimuth-elevation grid, and for every viewing direction we compare the bin-wise minimum range with the map's first-hit distance computed by raycasting. Furthermore, we apply a raycast consistency test that separates dynamic from static points. Finally, a spatial consistency validation step refines labels, producing static maps with lower residual dynamics and reduced over-removal. We evaluate our approach quantitatively and qualitatively on SemanticKITTI and a challenging custom dataset, and show consistent static mapping results.

\end{abstract}
\section{Introduction}

Static 3D mapping is essential for robot navigation, providing the geometric prior that supports reliable localization and planning~\cite{lim2021ral,sung2022isr}. A static map serves as the basis for simultaneous localization and mapping (SLAM), loop closure, and long-term mapping in changing environments~\cite{chen2022ral,lim2023rss}. A light detection and ranging (LiDAR) sensor is widely used to build 3D maps because of its accurate range and wide field of view, and it enables metric-scale mapping across large outdoor scenes. Over time, the map should remain consistent with subsequent LiDAR scans acquired from changing environments~\cite{maddern2015icra}.

However, dynamic scenes challenge this expectation. Moving objects leave residual returns on the 3D maps that do not belong to the static scene, while occlusions cause missing structures~\cite{wei2016ras}. Accumulated points of moving objects affect map consistency, degrading static mapping. The degraded static map ultimately has an adverse effect on robot navigation. Therefore, in static mapping, dynamic removal is necessary to reduce these side effects.

To address the problem of moving objects, numerous dynamic removal methods have been proposed~\cite{hornung2013ar,kim2020icra,mersch2022ral,schauer2018ral,oh2022ral}. Most static mapping methods handle dynamics by filtering measurements that are likely to be movable or by enforcing visibility consistency between the map and the current scan. These strategies often produce clean maps when the labels are reliable and the registration is accurate. 

Recently, various LiDAR sensors have been introduced, including mechanically spinning omnidirectional LiDAR (omni-LiDAR), solid-state LiDAR (solid-LiDAR), and flash-type LiDAR sensors. Thus, dynamic removal should be sensor-agnostic and robust to partial field of view (FoV)~\cite{lim2024iros}. However, existing methods implicitly assume omnidirectional coverage and uniform scan geometry, leading to residual structure or over-removal when the FoV is limited.

In this paper, we propose a novel dynamic removal module, Raymoval, which is a combination of the words Ray and Removal, for the static 3D mapping. Our contributions are threefold:

\begin{itemize}
  \item For each azimuth--elevation bin, we compare the bin-wise minimum range to the map's first-hit distance from raycasting, explicitly distinguishing dynamic and static points.
  \item We refine dynamic candidates with spatial consistency validation to recover static points and suppress fragments.
  \item Our method is sensor-agnostic and robustly removes dynamic points across diverse environments.
\end{itemize}

\section{Related Works}

Dynamic removal for static mapping has been extensively researched through several methods. The first is the visibility-consistency-based method. This method typically determines dynamic points by checking whether the query points are consistent with the visibility of the map points along each scan~\cite{xiao2015ijprs}. Classical probabilistic methods using occupancy grids are widely introduced, which update spaces as grid cells or voxels~\cite{hahnel2002iros,hornung2013ar}. Schauer and Nüchter~\cite{schauer2018ral} proposed a dynamic removal method by traversing an occupancy grid. Kim~and~Kim~\cite{kim2020icra} render both the accumulated map points and the query points as multiresolution range images and iteratively remove inconsistent points while reverting uncertain deletions near the occlusion boundaries to keep static points. These visibility pipelines are label-free and geometry-aware, but can be degraded when using partial FoV LiDAR sensors.

A complementary research adopts semantic or instance segmentation to remove dynamic objects while preserving static structure. Lim~\textit{et al.} proposed an efficient method to reject dynamic points comparing temporarily occupied bins~\cite{lim2021ral}. They detect temporal changes using the height difference of each bin based on the ground segmentation. Additionally, Lim~\textit{et al.} introduced instance-aware robust static mapping to handle dynamic scenes and imperfect segmentation~\cite{lim2023rss}. Other instance-aware methods combine geometric consistency with object-level hypotheses to robustly reject dynamic objects~\cite{chen2022ral,nunes2022ral}.

Learning-based moving object segmentation~(MOS) detects dynamic objects directly from LiDAR sequences. Chen~\textit{et al.}~\cite{chen2021ral} improved MOS by aggregating temporal changes across scans. And Mersch~\textit{et al.}~\cite{mersch2022ral} segment motion features with sparse 4D convolutions over a receding window and later maintain volumetric beliefs to keep a consistent static map while tracking dynamics. More recently, Jang~\textit{et al.} proposed a real-time MOS system that presents a voting-based method to achieve accurate dynamic removal and static recovery by emphasizing their spatiotemporal differences.

Recently, robustness for using heterogeneous LiDAR sensor setups has been important. Lim~\textit{et al.}~\cite{lim2024iros} proposed a dataset and benchmarks to provide point-wise MOS lables for point clouds captured by different types of LiDAR sensors. Also, dynamic removal should be sensor-agnostic and robust to partial FoV LiDAR sensors.

\section{Methodology}
\subsection{Overview}

\begin{figure}
    \centering
    \setlength{\abovecaptionskip}{3pt}
    \setlength{\belowcaptionskip}{0pt}
    \vspace{-6pt}
    \includegraphics[width=1.0\linewidth]{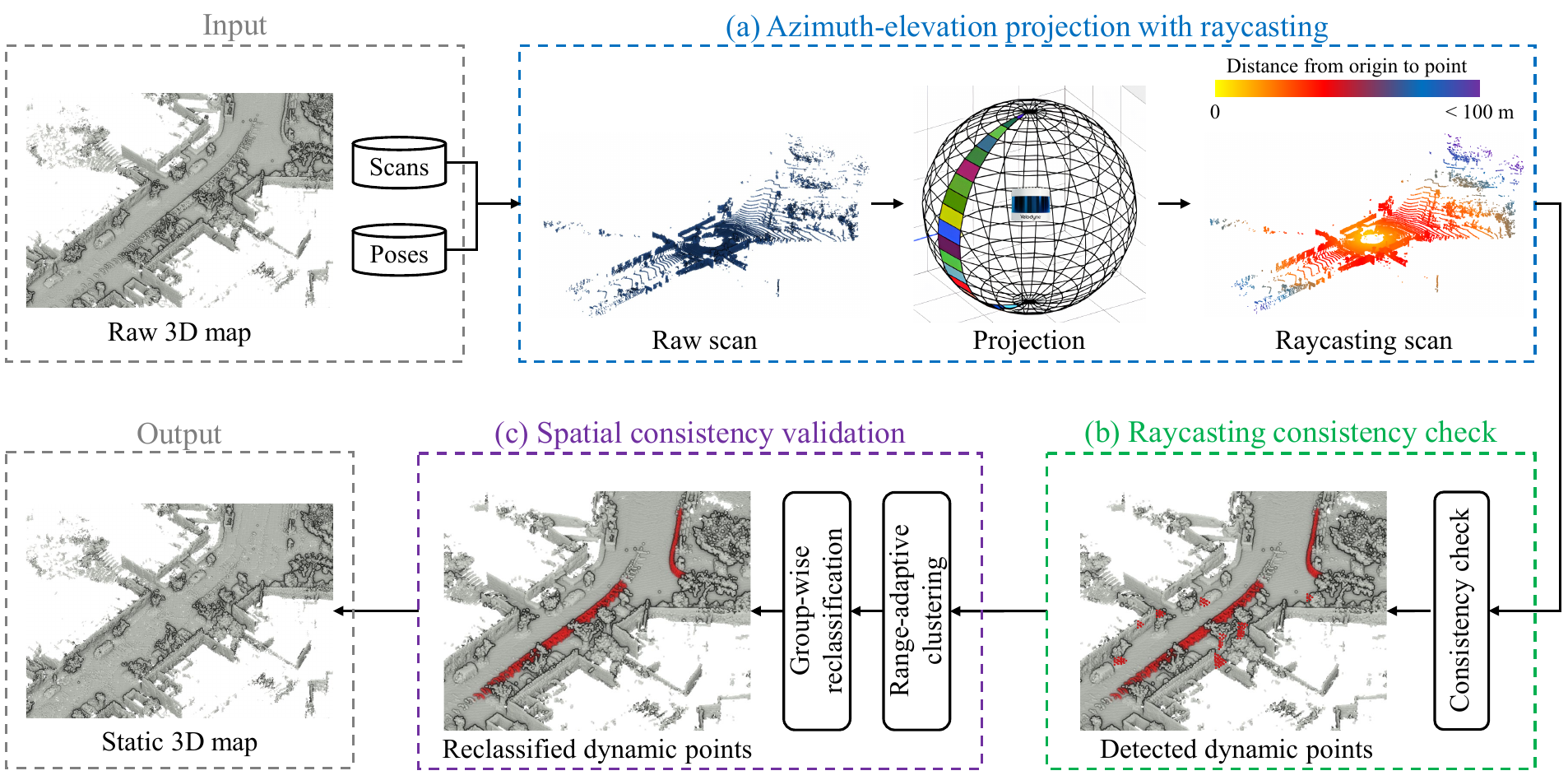}
    \caption{Overview of Raymoval pipeline. (a)~Each scan is projected to an azimuth--elevation grid to compute the raycasting, which stores the first-hit distance per viewing direction. (b)~A range-scaled consistency check compares scan ranges and raycasting map distances along the same visibility, determining dynamic points while keeping erase cases to prevent over-removal. (c)~A spatial consistency validation process performs range-adaptive clustering and group-wise reclassification to refine the labels~(best viewed in color).}
    \label{fig:overview}
\end{figure}

A pipeline of Raymoval is shown in Fig.~\ref{fig:overview}. Our approach aims to achieve robust dynamic removal using raycasting in a frame-wise manner. For each frame, we project the scan to azimuth--elevation grids as the pose, and obtain a raycasting. The raycasting is the first-hit distance along each viewing direction, so that every scan range can be compared along the same visibility.

The pipeline follows three steps. (a)~We project the scan to an azimuth--elevation grid and build the raycasting, using a small percentile over the neighbors to stabilize first-hit distances under partial FoV and occlusions. (b)~A range-scaled raycasting consistency check compares the scan ranges with the raycasting distances along the same viewing ray, determining foreground inconsistencies as dynamic points while keeping static points. (c)~A spatial consistency validation refines labels via range-adaptive clustering and group-wise reclassification with a dilated support test. This system helps to improve the robustness of the dynamic removal performance.

\subsection{Azimuth--Elevation Projection with Raycasting}

Our proposed method first converts a LiDAR scan in the sensor frame \(\mathcal{S}=\{\mathbf{p}_k^s\}\) into an azimuth--elevation (az--el) representation and constructs a raycasting, using the first-hit distance along each viewing direction. This section formalizes the grid construction and the first-hit distance obtained by raycasting from the current sensor pose.

\begin{figure}
    \centering
    \includegraphics[width=0.4\linewidth]{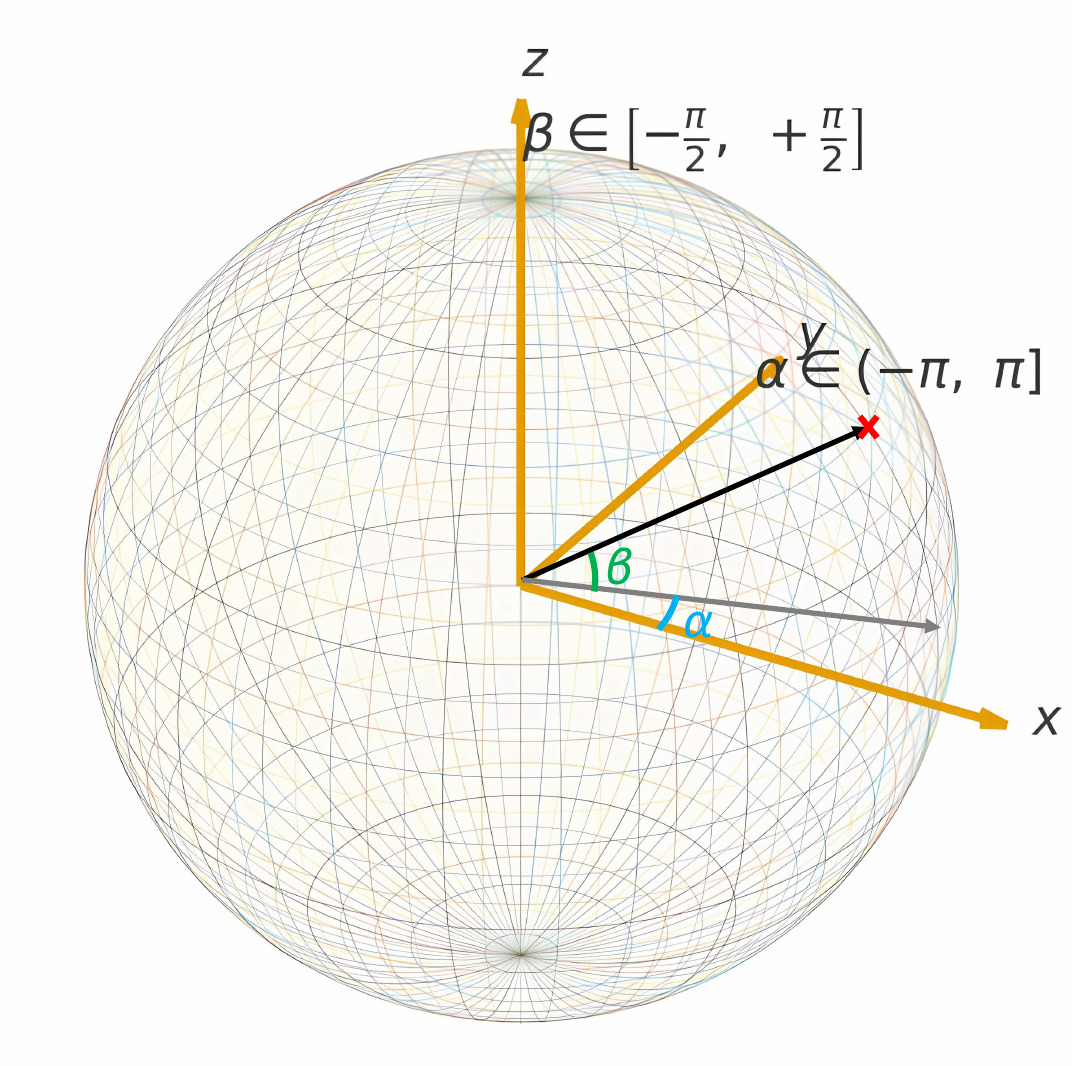}
    \caption{Visualization of azimuth--elevation representation. The raycasting stores the first-hit distance from the sensor origin along the bin's viewing direction to the nearest occupied point.}
    \label{fig:azel}
\end{figure}

We parameterize the scan directions by azimuth \(\alpha\) and elevation \(\beta\) as shown in Fig.~\ref{fig:azel}. For each scan point \(\mathbf{p}_k^s=[x_k,y_k,z_k]^\top\) with range \(r_k=\|\mathbf{p}_k^s\|\), azimuth $\alpha$ and elevation $\beta$ are defined as follows:

\begin{equation}
\label{eq:angles}
\alpha_k=\operatorname{arctan2}(y_k,x_k),\qquad
\beta_k=\arcsin\!\Big(\frac{z_k}{r_k}\Big).
\end{equation}
The operator \(\operatorname{arctan2}(y,x)\) denotes the arctangent of two arguments returning the principal angle of \((x,y)\) with the correct quadrant. The azimuth is wrapped to \((-\pi,\pi]\); elevation is limited by the vertical FoV of LiDAR to \([\beta_{\min},\beta_{\max}]\). The angles are then discretized into uniform bins \(N_{\mathrm{az}}\) and \(N_{\mathrm{el}}\). Let the step sizes be:

\begin{equation}
\delta_\alpha=\frac{2\pi}{N_{\mathrm{az}}},\qquad
\delta_\beta=\frac{\beta_{\max}-\beta_{\min}}{N_{\mathrm{el}}}.
\end{equation}

Each point is assigned to a bin \((i_k,j_k)\) and the corresponding angle of the center of the bin $\bar\alpha_i$ and $\bar\beta_j$ are defined as follows:

\begin{equation}
i_k=\Big\lfloor \frac{\alpha_k+\pi}{\delta_\alpha}\Big\rfloor,\qquad
j_k=\Big\lfloor \frac{\beta_k-\beta_{\min}}{\delta_\beta}\Big\rfloor.
\end{equation}

\begin{equation}
\bar\alpha_i=-\pi+\Big(i+\tfrac12\Big)\delta_\alpha,\qquad
\bar\beta_j=\beta_{\min}+\Big(j+\tfrac12\Big)\delta_\beta.
\end{equation}

For bin \((i,j)\), the sensor-frame direction vector to the corresponding bin is as $\mathbf{u}_{ij}$:

\begin{equation}
\mathbf{u}_{ij}=
\begin{bmatrix}
\cos\bar\alpha_i\,\cos\bar\beta_j\\
\sin\bar\alpha_i\,\cos\bar\beta_j\\
\sin\bar\beta_j
\end{bmatrix},
\qquad \|\mathbf{u}_{ij}\|_2=1.
\end{equation}

To compare the obtained raycasting with the map in the same visibility, we express the direction vector of each bin in the world frame where the prior scene \(\mathcal{M}\) is defined. Let \(w\) and \(s\) denote the world and sensor frames, respectively, and let \(\mathbf{T}_s^w=[\,\mathbf{R}_s^w\mid \mathbf{t}_s^w\,]\) coordinates from \(s\) to \(w\). Then, the raycasting for bin \((i,j)\) in the world frame is defined as follows:

\begin{equation}
\mathbf{o}^w=\mathbf{t}_s^w,\qquad
\mathbf{d}_{ij}^w=\mathbf{R}_s^w\,\mathbf{u}_{ij}^s.
\end{equation}

\begin{equation}
r_{\mathrm{cast}}[i,j]
=\inf\{\lambda\in(0,r_{\max}]:\ \mathbf{o}^w+\lambda\,\mathbf{d}_{ij}^w\in\partial\mathcal{M}\}.
\end{equation}

\noindent
with \(r_{\mathrm{cast}}[i,j]=+\infty\) if no intersection occurs within \(r_{\max}\).
Here, \(\mathbf{R}_s^w\in\mathrm{SO}(3)\) and \(\mathbf{t}_s^w\in\mathbb{R}^3\) are the rotation and translation of the pose \(\mathbf{T}_s^w\) (sensor \(\to\) world). \(\mathbf{o}^w=\mathbf{t}_s^w\) is the origin of the scan in the world frame and \(\mathbf{d}_{ij}^w=\mathbf{R}_s^w\,\mathbf{u}_{ij}^s\) is the corresponding direction vector. \(r_{\max}>0\) is a truncated range to reject outliers and to avoid large computations. \(\mathcal{M}\subset\mathbb{R}^3\) denotes a prior voxelized map and \(\partial\mathcal{M}\) is its occupied boundary. The array \(r_{\mathrm{cast}}[i,j]\) forms the raycasting.

\subsection{Raycasting Consistency Check}

We compare each scan with the map along the same viewing ray. Using a single bin to detect dynamic points can be degraded under partial FoV, occlusions, and LiDAR noise. To mitigate these effects, we aggregate first-hit distances over a small az--el neighborhood and take a small percentile as the expected map distance. This stabilizes the comparison and reduces the sensitivity to outliers.

We use a Chebyshev neighborhood of radius $r_n\!\in\!\mathbb{N}$ around the bin $(i,j)$ to obtain a robust map reference. Thus, we keep only those directions where the ray actually meets an occupied voxel, and directions without a hit are ignored. With the square window $\mathcal{W}_{ij}(r_n)=\{(u,v):\|(u-i,v-j)\|_\infty\le r_n\}$, we can form a robust map reference by a lower quantile:

\begin{equation}
\label{eq:win_hits}
\mathcal{D}_{ij}=\big\{\,r_{\mathrm{cast}}[u,v]\;:\;(u,v)\in\mathcal{W}_{ij}(r_n),\ r_{\mathrm{cast}}[u,v]<\infty\,\big\}.
\end{equation}

A robust map reference is then obtained by a lower quantile ($q\in(0,1)$) for robustness:
\begin{equation}
\label{eq:dmap}
d^{\mathrm{map}}_{ij}=\Perc_{q}\!\left(\mathcal{D}_{ij}\right).
\end{equation}

Let $\mathcal{S}_{ij}$ be the set of scan points assigned to bin $(i,j)$. For the scan, we take the nearest point in bin $(i,j)$ and compare it with $d^{\mathrm{map}}_{ij}$ along the same visibility:
\begin{equation}
\label{eq:dscan}
d^{\mathrm{scan}}_{ij}=
\begin{cases}
\min_{\mathbf{p}\in\mathcal{S}_{ij}}\|\mathbf{p}\|_2, & \mathcal{S}_{ij}\neq\emptyset,\\
\infty, & \mathcal{S}_{ij}=\emptyset.
\end{cases}
\end{equation}

Finally, we decide the label with a simple rule. If the scan lies sufficiently in front of the map reference, the points are dynamic, otherwise static.
\begin{equation}
e_{ij}=d^{\mathrm{map}}_{ij}-d^{\mathrm{scan}}_{ij},\qquad
\ell_{ij}=
\begin{cases}
\textsc{dynamic}, & e_{ij} > \tau_0 + \tau_1\,d^{\mathrm{scan}}_{ij},\\
\textsc{static},  & \text{otherwise.}
\end{cases}
\end{equation}
Here, $\Perc_{q}$ provides a stable map reference near depth discontinuities, and the range-adaptive margin $\tau(d)$ (linear in distance) mitigates the effects of outliers.

\subsection{Spatial Consistency Validation}

The raycasting consistency check may cause objects to fragment and over-remove boundary points near depth discontinuities. Therefore, we refine the dynamic points using spatial consistency validation. This consists of range-adaptive clustering and group-wise reclassification.

We first cluster dynamic points by Euclidean distance. For a cluster $C_k$, let $r_k=\operatorname{median}_{\mathbf{x}\in C_k}\|\mathbf{x}\|_2$ and define the horizontal diameter of its bounding box.

\begin{equation}
\operatorname{diam}(C_k)=\sqrt{(\Delta x)^2+(\Delta y)^2}.
\end{equation}

where $\Delta x=x_{\max}-x_{\min}$ and $\Delta y=y_{\max}-y_{\min}$. We keep clusters only if they are large and wide enough in that range:
\begin{equation}
\label{eq:range_screen}
|C_k|\ \ge\ m(r_k),\qquad \operatorname{diam}(C_k)\ \ge\ d(r_k),
\end{equation}
where $m(\cdot)$ and $d(\cdot)$ are range-adaptive functions. This removes tiny or far fragments that are unlikely to be real objects.

Next, we perform group-wise reclassification against the static map with a small dilation. Let $\mathcal{M}$ be the static map and $\mathcal{M}^{\oplus}$ its Chebyshev dilation by $\delta_v$ voxels of size $v$. For a set $A$ of points, the coverage is the ratio of points that lie inside $\mathcal{M}^{\oplus}$:
\begin{equation}
\label{eq:cov_def}
\operatorname{coverage}(A)=\frac{\big|\{\mathbf{x}\in A:\ \mathbf{x}\in\mathcal{M}^{\oplus}\}\big|}{|A|}.
\end{equation}

We then merge the clusters that are adjacent in the voxel grid into a group $G$, so that the split pieces of the same object are handled together. For each group, we compute the coverage $\operatorname{coverage}(G)$ by \eqref{eq:cov_def}, and the minimum side of its bounding box
$\operatorname{thin}(G)=\min\{\Delta x,\Delta y,\Delta z\}$, where $\Delta z=z_{\max}-z_{\min}$. We set $G$ back to static if it overlaps the map enough or if it is a thin structure with sufficient support:

\begin{equation}
\operatorname{coverage}(G)\ \ge\ \theta_{\mathrm{coverage}}
\quad\text{or}\quad
\big(\operatorname{coverage}(G)\ \ge\ \theta_{\mathrm{edge}}\ \wedge\ \operatorname{thin}(G)\ \le\ \theta_{\mathrm{thin}}\big).
\end{equation}

For groups that remain dynamic, we keep only the unsupported points of each cluster. Thus, small residual groups are removed to avoid fragments. if the remaining size is below $m_{\min}$ or its horizontal diameter is below $d_{\min}$, the group is reclassified to static points. This step restores static surfaces near depth edges and keeps only consistent dynamic objects.

\section{Experiments}

\subsection{Experimental Setup}

\begin{figure}
    \centering
    \includegraphics[width=0.8\linewidth]{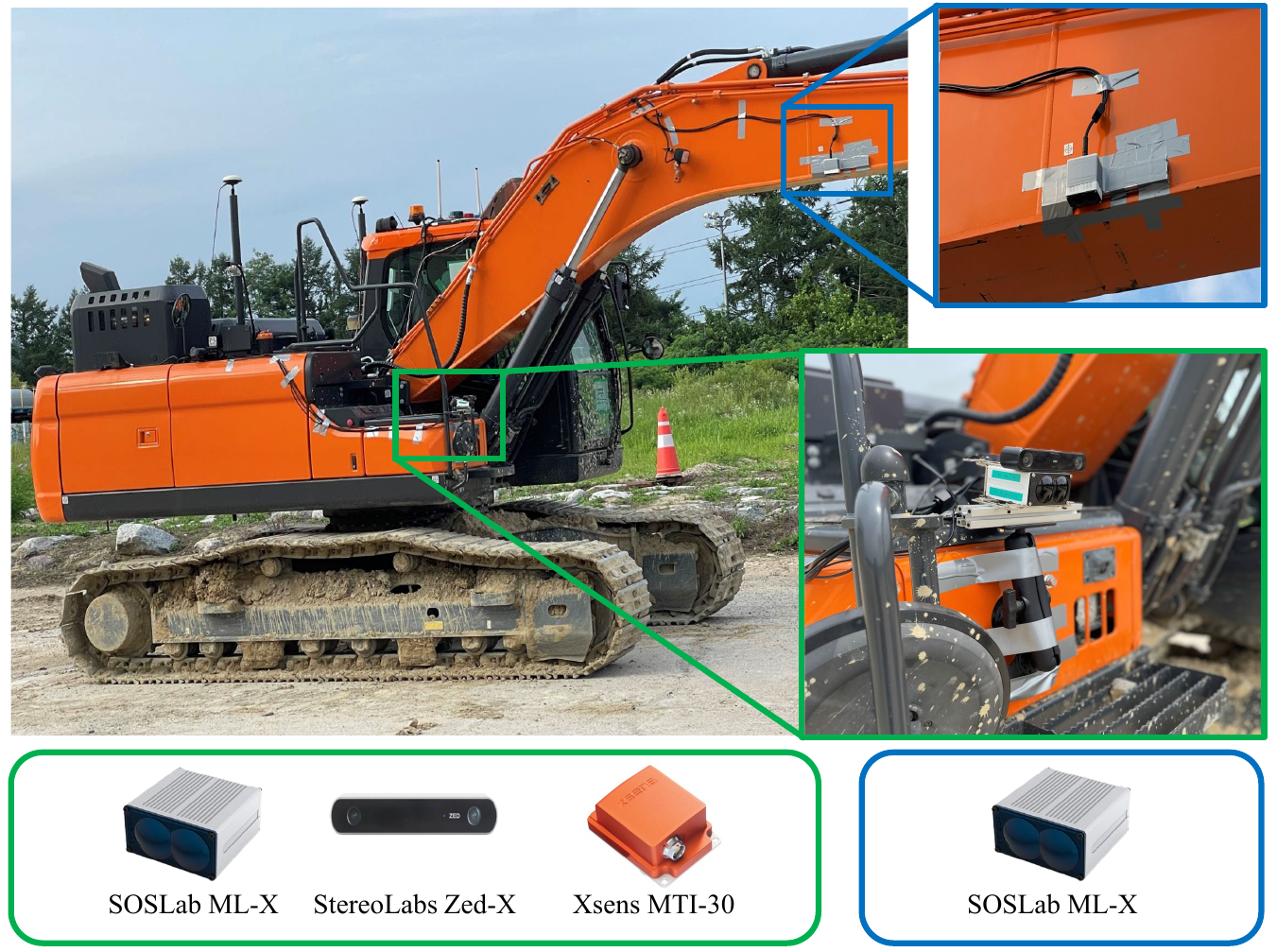}
    \caption{A picture of the excavator platform that acquired our custom dataset. A rigid sensors suite~(SOSLab ML-X LiDAR, StereoLabs Zed-X camera, and Xsens MTI-30 IMU) is mounted on the excavator body, and an additional downward-facing SOSLab ML-X LiDAR is mounted on the boom to detect the ground. Data were acquired in active construction sites while the machine was moving and excavating.}
    \label{fig:dataset}
\end{figure}

To evaluate the dynamic removal performance of Raymoval in various environments, we evaluated our proposed method on the SemanticKITTI dataset~\cite{behley2019kitti} and on a custom dataset collected with the excavator platform as shown in~Fig.~\ref{fig:dataset}.

SemanticKITTI provides omni-LiDAR scans, while our challenging custom dataset was collected with a solid-LiDAR. Evaluating both settings demonstrates that our proposed method is sensor-agnostic and works under full and partial FoV.

To estimate the poses in our challenging custom dataset, we used LVI-Q~\cite{kevin2025ral}, which is an algorithm to robustly estimate the robot pose using vision, IMU, LiDAR, and kinematics information.

\textbf{SemanticKITTI Dataset.}
We adopt the ERASOR benchmark~\cite{lim2021ral} on SemanticKITTI and follow the same LiDAR-only protocol. The subset contains five segments where moving objects frequently appear: Seq.~00 (4390–4530), 01 (150–250), 02 (860–950), 05 (2350–2670), and 07 (630–820). We use the provided odometry poses so that the evaluation isolates the effect of dynamic removal on static map building.

\textbf{Challenging Custom Dataset.}
We also evaluated data collected at active construction sites using our excavator platform (Fig.~\ref{fig:dataset}). A rigid sensor set is mounted on the body, and an additional downward-facing solid-LiDAR is mounted on the boom to better observe the ground. This setting stresses partial FoV, occlusions, and rapidly changing scenes. During rotation and excavation, the platform has a lot of large motion and vibration, and the LiDAR frequently observes the excavator body and boom; these parts move with respect to the world, producing dynamic environments and self-occlusions in the scans. The solid-LiDAR further limits coverage and has non-uniform scan geometry, creating strong partial-FoV and ego-induced dynamics. The ground truth point cloud is obtained by Trimble 3D laser scanner after taking the dataset.

\textbf{Evaluation Metrics}
To quantitatively evaluate dynamic removal performance, we utilize \textit{Preservation rate (PR)}, \textit{Rejection rate (RR)}, and \textit{F1 score}, as proposed by Lim~\textit{et al.}~\cite{lim2021ral}:
\begin{itemize}
\item PR $=\dfrac{\footnotesize{\#\ \text{of preserved static voxels}}}{\footnotesize{\#\ \text{of total static voxels on the naively accumulated map}}}$
\item RR $=1-\dfrac{\footnotesize{\#\ \text{of remaining dynamic voxels}}}{\footnotesize{\#\ \text{of total dynamic voxels on the naively accumulated map}}}$
\item $F_{1}=\dfrac{2\,\text{PR}\cdot\text{RR}}{\text{PR}+\text{RR}}$
\end{itemize}

\textbf{Hyperparameter Configuration}

All experiments were conducted using the same parameter settings unless otherwise noted. The key hyperparameters related to raycasting, voxelization, and clustering are summarized in Table~\ref{tab:params}. Raymoval employs two distance thresholds, $\tau_0$ and $\tau_1$ for dynamic–static discrimination. The resolution of azimuth–-elevation grid $(N_\mathrm{az}, N_\mathrm{el}) = (720, 450)$ covers the entire FoV. A voxel size of $0.2$\,m is used for both preprocessing and map integration. Neighborhood filtering with $(r_n, q) = (1, 0.9)$ stabilizes the raycasting reference under occlusion and partial FoV. All parameters were kept fixed across the datasets to demonstrate the sensor-agnostic characteristic of Raymoval.

\begin{table}[t!]
\centering
\scriptsize
\caption{Hyperparameters used in all experiments.}
\label{tab:params}
\renewcommand{\arraystretch}{1.1}
\setlength{\tabcolsep}{4pt}
\begin{tabular}{lcl}
\hline
& \textbf{Parameter} & \textbf{Value} \\
\hline
& $(N_\mathrm{az}$, $N_\mathrm{el})$ & (720, 450) divisions \\
& Range limit & 60.0\,m \\
& Distance thresholds & $\tau_0 = 0.30$ m,\quad $\tau_1 = 0.35$ m \\
& Neighborhood filter & $r_n=1$ bin,\quad percentile $q=0.9$ \\
& Voxel size & 0.2\,m \\
\hline
\end{tabular}
\vspace{-3pt}
\end{table}

\subsection{Results}

\begin{table}[h!]
\centering
\caption{Quantitative comparison on the SemanticKITTI benchmark (PR: Preservation Rate, RR: Rejection Rate). The best performance is highlighted in bold, and the second-best result is underlined.}
\label{tab:kitti_quant_cmp}
\setlength{\tabcolsep}{6pt}
\renewcommand{\arraystretch}{1.15}
{
\tiny
\begin{tabular}{ccccc}
\Xhline{1.2pt}
\noalign{\vskip 2pt}
\hline
Seq. & Method & PR [\%] & RR [\%] & $F_{1}$ score\\
\hline
\multirow{5}{*}{00}
& OctoMap - 0.2~\cite{hornung2013ar} & 34.568 & \textbf{99.979} & 0.514 \\
& Peopleremover~\cite{schauer2018ral} & 37.523 & 89.116 & 0.528\\
& Removert - \texttt{RM3+RV1}~\cite{kim2020icra} & 86.829 & 90.617 & 0.887 \\
& ERASOR~\cite{lim2021ral} & \underline{93.980} & \underline{97.081} & \textbf{0.955} \\
& Raymoval (proposed) & \textbf{94.046} & 90.428 & \underline{0.922} \\
\hline
\multirow{5}{*}{01}
& OctoMap - 0.2~\cite{hornung2013ar} & 20.777 & \textbf{99.863} & 0.344 \\
& Peopleremover~\cite{schauer2018ral} & 36.349 & 93.116 & 0.523 \\
& Removert - \texttt{RM3+RV1}~\cite{kim2020icra} & \textbf{95.815} & 57.077 & 0.715 \\
& ERASOR~\cite{lim2021ral} & 91.487 & \underline{95.383} & \textbf{0.934} \\
& Raymoval (proposed) & \underline{91.854} & 92.051 & \underline{0.920} \\
\hline
\multirow{5}{*}{02}
& OctoMap - 0.2~\cite{hornung2013ar} & 23.746 & \textbf{99.792} & 0.384 \\
& Peopleremover~\cite{schauer2018ral} & 29.037 & 94.527 & 0.444 \\
& Removert - \texttt{RM3+RV1}~\cite{kim2020icra} & 83.293 & 88.371 & 0.858 \\
& ERASOR~\cite{lim2021ral} & \underline{87.731} & \underline{97.008} & \underline{0.921} \\
& Raymoval (proposed) & \textbf{95.144} & 93.181 & \textbf{0.942} \\
\hline
\multirow{5}{*}{05}
& OctoMap - 0.2~\cite{hornung2013ar} & 33.904 & \textbf{99.882} & 0.506 \\
& Peopleremover~\cite{schauer2018ral} & 38.495 & 90.631 & 0.540 \\
& Removert - \texttt{RM3+RV1}~\cite{kim2020icra} & 88.170 & 79.981 & 0.839 \\
& ERASOR~\cite{lim2021ral} & \underline{88.730} & \underline{98.262} & \underline{0.933} \\
& Raymoval (proposed) & \textbf{93.394} & 95.636 & \textbf{0.945} \\
\hline
\multirow{5}{*}{07}
& OctoMap - 0.2~\cite{hornung2013ar} & 38.183 & \textbf{99.565} & 0.552 \\
& Peopleremover~\cite{schauer2018ral} & 34.772 & 91.983 & 0.505 \\
& Removert - \texttt{RM3+RV1}~\cite{kim2020icra} & 82.038 & 95.504 & 0.883 \\
& ERASOR~\cite{lim2021ral} & \underline{90.624} & \underline{99.271} & \textbf{0.948} \\
& Raymoval (proposed) & \textbf{91.645} & 91.534 & \underline{0.916} \\
\hline
\multirow{5}{*}{Average}
& OctoMap - 0.2~\cite{hornung2013ar} & 30.236 & \textbf{99.816} & 0.460 \\
& Peopleremover~\cite{schauer2018ral} & 35.235 & 91.875 & 0.508 \\
& Removert - \texttt{RM3+RV1}~\cite{kim2020icra} & 50.700 & 82.310 & 0.836 \\
& ERASOR~\cite{lim2021ral} & \underline{90.510} & \underline{97.401} & \textbf{0.938} \\
& Raymoval (proposed) & \textbf{93.217} & 92.566 & \underline{0.927} \\
\hline
\noalign{\vskip 2pt}
\Xhline{1.2pt}
\end{tabular}
}
\end{table}

\begin{figure}[h!]
    \centering
    \includegraphics[width=0.95\linewidth]{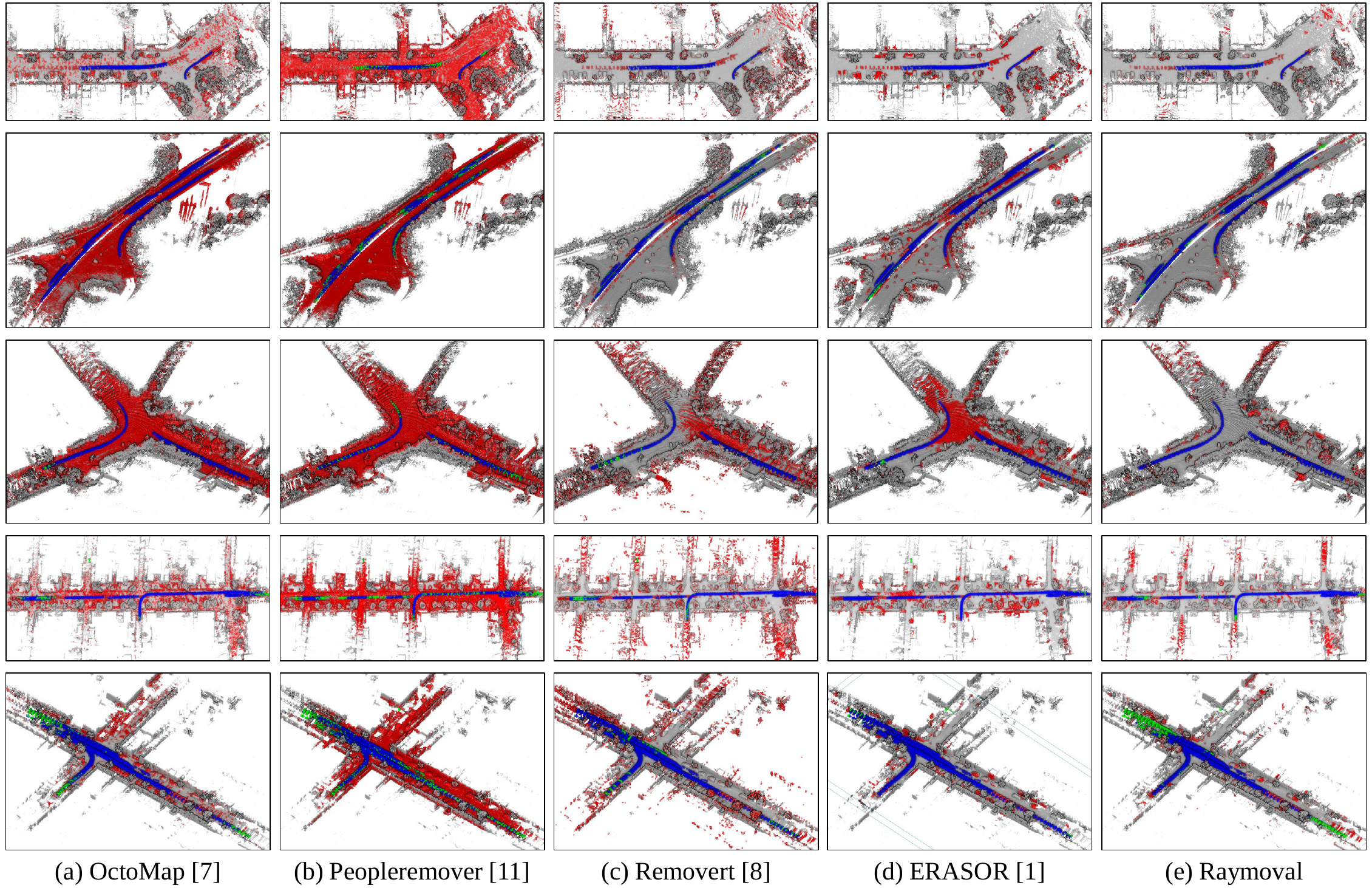}
    \caption{(a)-(e) Qualitative comparison with the state-of-the-art dynamic removal approaches on 00, 01, 02, 05, 07 sequences in the SemanticKITTI dataset~(from top to bottom). Green, red, and blue points indicate true positive~(TP), false positive~(FP), and false negative~(FN), respectively~(best viewed in color).}
    \label{fig:viz_map}
\end{figure}

Table~\ref{tab:kitti_quant_cmp} summarizes the five SemanticKITTI sequences. Raymoval shows equal or substantially better performance with other state-of-the-art~(SOTA) algorithms. Across all sequences, it ranks first or second in $F_{1}$, while showing a high rejection rate (RR) and improving preservation rate (PR).

As shown in Fig.~\ref{fig:viz_map}, we demonstrate that our Raymoval accurately removes only dynamic objects, without over-removal for static points. While the state-of-the-art dynamic removal approaches showed over-removal results, our Raymoval successfully remove the dynamic points and keep the static points.

\begin{figure}[h!]
    \centering
    \includegraphics[width=0.95\linewidth]{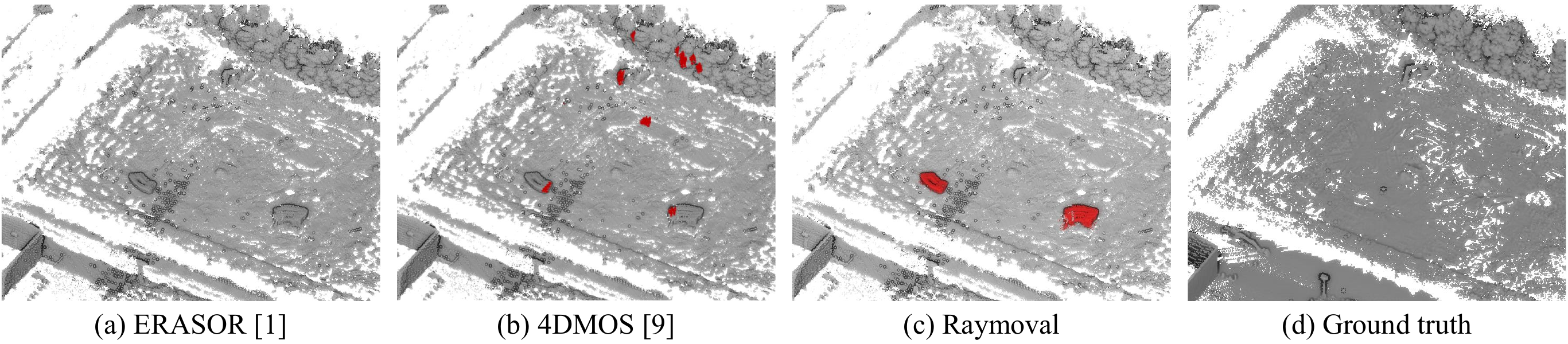}
    \caption{(a)-(c) Qualitative comparison with the state-of-the-art dynamic removal approaches on our challenging custom dataset. Red points indicate dynamic points~(best viewed in color).}
    \label{fig:hd_exp}
\end{figure}

The qualitative results on our challenging dataset in construction site are shown in Fig.~\ref{fig:hd_exp}. They further demonstrate that our proposed approach robustly removes moving objects under partial FoV. The dataset includes self-occlusion by the boom and body, unstructured ground. In these scenes, Raymoval suppresses over-removal, successfully removes the excavator's body and boom, and keeps the ground geometry consistent over time.

Table~\ref{tab:runtime} summarizes the runtime of Raymoval. The raycasting cache construction dominates the total computation, accounting for about 93\% of the processing time, while the scan binning, classification, and no-return evidence steps are negligible. The average of total runtime in SemanticKITTI dataset is 93.8\,ms per scan, corresponding to approximately 10.6\,Hz on a single CPU thread.

\begin{table}[t!]
\centering
\footnotesize
\caption{Average runtime of Raymoval per a scan in the SemanticKITTI dataset on Intel(R) Core(TM) i9-13900 CPU.}
\label{tab:runtime}
\renewcommand{\arraystretch}{1.1}
\setlength{\tabcolsep}{8pt}
\begin{tabular}{lcc}
\hline
\textbf{Stage} & \textbf{Time [ms]} & \textbf{Percentage [\%]} \\
\hline
Scan binning & 1.05 & 1.1 \\
Raycasting cache & 87.11 & 92.9 \\
Classification & 5.05 & 5.4 \\
No-return evidence & 0.55 & 0.6 \\
\hline
\textbf{Total} & \textbf{93.83} & \textbf{100.0} \\
\hline
\end{tabular}
\vspace{-3pt}
\end{table}

\section{Conclusion}

In this paper, we propose Raymoval, a raycasting-based dynamic removal module for static 3D mapping. Each scan is projected to azimuth–elevation bins, and a lower-quantile raycast provides a stable map reference per viewing direction. A simple range-adaptive test separates the dynamic objects from the static structure, and a spatial consistency validation step recovers boundary points and misclassified points. Unlike voxel-averaging or height-map approaches, the raycasting formulation preserves per-ray visibility consistency, enabling precise dynamic detection under partial FoV and reducing over-removal near depth discontinuities.

In SemanticKITTI and our challenging construction-site dataset with solid-LiDAR, our method robustly preserves static points while removing dynamic points under both full and partial FoV. The results show reduced over-removal near depth edges and consistent maps across diverse scenes and sensors.

As an extension, the same evidence used for removal can be aggregated to update long-term map changes at once. A batch change update over voxels could be applied to structural additions and deletions simultaneously, enabling a unified “clean-and-update” static mapping stage.

\section{Acknowledgement}
This research was supported in part by the KAIST Quantum+X Convergence R\&D Project and in part by HD Hyundai XiteSolution.
The students are supported by BK21 FOUR.

%
%
%
%

\end{document}